\documentclass[11pt,a4paper]{article}

\usepackage{url}
\usepackage{soul}
\usepackage[hidelinks]{hyperref}
\usepackage[utf8]{inputenc}
\usepackage[small]{caption}
\usepackage{graphicx}
\usepackage{booktabs}
\usepackage{algorithm}
\usepackage{algorithmic}
\usepackage{comment}
\usepackage{amsmath}
\usepackage{amssymb}
\usepackage{color}
\usepackage{multirow}
\usepackage{makecell}
\usepackage[hyperref]{emnlp-ijcnlp-2019}
\usepackage{times}
\usepackage{latexsym}

\aclfinalcopy 

\title{Open Event Extraction from Online Text using a Generative Adversarial Network}

\author{Rui Wang$^{\dag}$ \ \ \ \ \   Deyu Zhou$^{*\dag}$ \ \ \ \ \ Yulan He$^{\S}$\\
  $^{\dag}$School of Computer Science and Engineering, Key Laboratory of Computer Network\\
  and Information Integration, Ministry of Education, Southeast University, China \\
  $\S$Department of Computer Science, University of Warwick, UK \\
  \{rui\_wang, d.zhou\}@seu.edu.cn,  yulan.he@warwick.ac.uk\\}

\date{}

\begin{document}
\maketitle

\begin{abstract}
To extract the structured representations of open-domain events, Bayesian graphical models have made some progress.  However, these approaches typically assume that all words in a document are generated from a single event. While this may be true for short text such as tweets, such an assumption does not generally hold for long text such as news articles. Moreover, Bayesian graphical models often rely on Gibbs sampling for parameter inference which may take long time to converge. To address these limitations, we propose an event extraction model based on Generative Adversarial Nets, called Adversarial-neural Event Model (AEM). AEM models an event with a Dirichlet prior and uses a generator network to capture the patterns underlying latent events. A discriminator is used to distinguish documents reconstructed from the latent events and the original documents. A byproduct of the discriminator is that the features generated by the learned discriminator network allow the visualization of the extracted events. Our model has been evaluated on two Twitter datasets and a news article dataset. Experimental results show that our model outperforms the baseline approaches on all the datasets, with more significant improvements observed on the news article dataset where an increase of 15\% is observed in F-measure. 
\end{abstract}

\section{Introduction}

With the increasing popularity of the Internet, online texts provided by social media platform (e.g. Twitter) and news media sites (e.g. Google news) have become important sources of real-world events. Therefore, it is crucial to automatically extract events from online texts.

Due to the high variety of events discussed online and the difficulty in obtaining annotated data for training, traditional template-based or supervised learning approaches for event extraction are no longer applicable in dealing with online texts. Nevertheless, newsworthy events are often discussed by many tweets or online news articles. Therefore, the same event could be mentioned by a high volume of redundant tweets or news articles. This property inspires the research community to devise clustering-based models \cite{popescu2011extracting,abdelhaq2013eventweet,xia2015new} to discover new or previously unidentified events without extracting structured representations. 

 {\color{black}To extract structured representations of events such as \emph{who} did \emph{what}, \emph{when}, \emph{where} and \emph{why}, Bayesian approaches have made some progress.}  {\color{black}Assuming that each document is assigned to a single event, which is modeled as a joint distribution over the named entities, the date and the location of the event, and the event-related keywords,} Zhou \emph{et al.}~\shortcite{zhou2014simple} proposed an unsupervised Latent Event Model (LEM) for open-domain event extraction. To address the limitation that LEM requires the number of events to be pre-set, Zhou \emph{et al.}~\shortcite{zhou2017event} further proposed the Dirichlet Process Event Mixture Model (DPEMM) in which the number of events can be learned automatically from data. However, both LEM and DPEMM have two limitations: (1) they assume that all words in a document are generated from a single event which can be represented by a quadruple $<$entity, location, keyword, date$>$. However, long texts such news articles often describe multiple events which clearly violates this assumption; 
 {\color{black}(2) During the inference process of both approaches, the Gibbs sampler needs to compute the conditional posterior distribution and assigns an event for each document. This is time consuming and takes long time to converge.}

To deal with these limitations, in this paper, we propose the Adversarial-neural Event Model (AEM) based on adversarial training for open-domain event extraction. The principle idea is to use a generator network to learn the projection function between the document-event distribution and four event related word distributions (entity distribution, location distribution, keyword distribution and date distribution). Instead of providing an analytic approximation, AEM uses a discriminator network to discriminate between the reconstructed documents from latent events and the original input documents. This essentially helps the generator to construct a more realistic document from a random noise drawn from a Dirichlet distribution. Due to the flexibility of neural networks, the generator is capable of learning complicated nonlinear distributions. And the supervision signal provided by the discriminator will help generator to capture the event-related patterns. Furthermore, the discriminator also provides low-dimensional discriminative features which can be used to visualize documents and events. 

The main contributions of the paper are summarized below:
\begin{itemize}
	\item {\color{black}We propose a novel Adversarial-neural Event Model (AEM), which is, to the best of our knowledge, the first attempt of using adversarial training for open-domain event extraction. }
	\item {\color{black}Unlike existing Bayesian graphical modeling approaches, AEM is able to extract events from different text sources (short and long). And a significant improvement on computational efficiency is also observed.}
	\item {\color{black}Experimental results on three datasets show that AEM outperforms the baselines in terms of accuracy, recall and F-measure. In addition, the results show the strength of AEM in visualizing events.} 
\end{itemize}

\section{Related Work}

{\color{black}Our work is related to two lines of research, event extraction and Generative Adversarial Nets.}

\subsection*{Event Extraction}

Recently there has been much interest in event extraction from online texts, and approaches could be categorized as domain-specific and open-domain event extraction.

Domain-specific event extraction often focuses on the specific types of events (e.g. sports events or city events). Panem \emph{et al.}~\shortcite{panem2014structured} devised a novel algorithm to extract attribute-value pairs and mapped them to manually generated schemes for extracting the natural disaster events. Similarly, to extract the city-traffic related event, Anantharam \emph{et al.}~\shortcite{anantharam2015extracting} viewed the task as a sequential tagging problem and proposed an approach based on the conditional random fields. {\color{black} Zhang~\shortcite{zhang2018event} proposed an event extraction approach based on imitation learning, especially on inverse reinforcement learning. }

Open-domain event extraction aims to extract events without limiting the specific types of events. To analyze individual messages and induce a canonical value for each event, Benson \emph{et al.}~\shortcite{benson2011event} proposed an approach based on a structured graphical model. By representing an event with a binary tuple which is constituted by a named entity and a date, Ritter \emph{et al.}~\shortcite{ritter2012open} employed some statistic to measure the strength of associations between a named entity and a date. The proposed system relies on a supervised labeler trained on annotated data. In \cite{abdelhaq2013eventweet}, Abdelhaq \emph{et al.} developed a real-time event extraction system called EvenTweet, and each event is represented as a triple constituted by time, location and keywords. To extract more information, Wang \emph{el al.}~\shortcite{wang2015seeft} developed a system employing the links in tweets and combing tweets with linked articles to identify events. Xia \emph{el al.}~\shortcite{xia2015new} combined texts with the location information to detect the events with low spatial and temporal deviations. Zhou \emph{et al.}~\shortcite{zhou2014simple,zhou2017event} represented event as a quadruple and proposed two Bayesian models to extract events from tweets.

\subsection*{Generative Adversarial Nets}

As a neural-based generative model, Generative Adversarial Nets \cite{goodfellow2014generative} have been extensively researched in natural language processing (NLP) community. 

\begin{figure*}[!htbp]
\centering
\includegraphics[
  width=0.9\textwidth,
  keepaspectratio]
{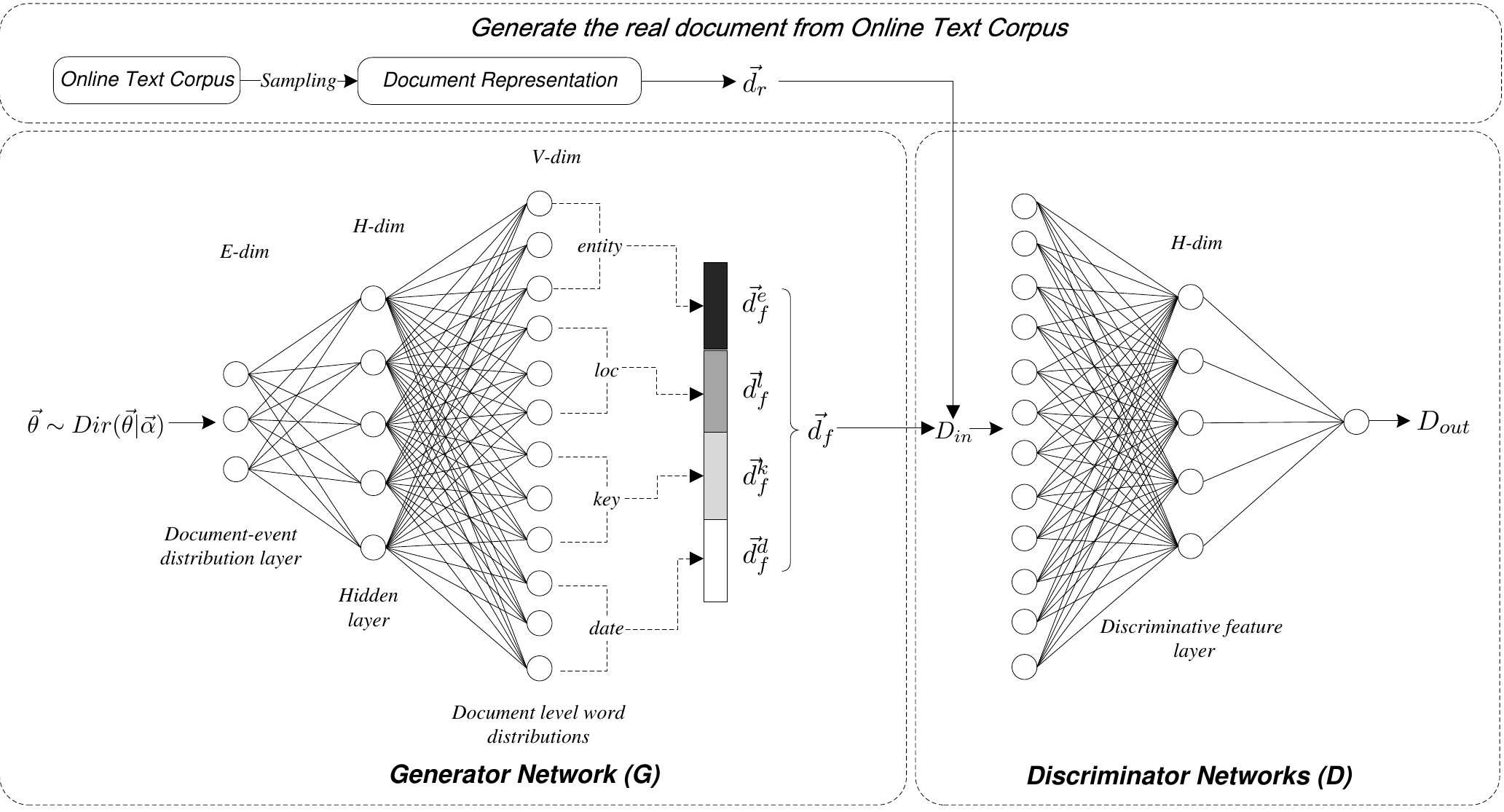}
\caption{The framework of the Adverarial-neural Event Model (AEM), {\color{black}and $\vec d_{f}^{e}$, $\vec d_{f}^{l}$, $\vec d_{f}^{k}$ and $\vec d_{f}^{d}$} denote the generated entity distribution, location distribution, keyword distribution and date distribution corresponding to event distribution $\vec \theta$.}
\label{fig:aem_framework}
\end{figure*}

For text generation, the sequence generative adversarial network (SeqGAN) proposed in \cite{yu2017seqgan} incorporated a policy gradient strategy to optimize the generation process. Based on the policy gradient, Lin \emph{et al.}~\shortcite{lin2017adversarial} proposed RankGAN to capture the rich structures of language by ranking and analyzing a collection of human-written and machine-written sentences. To overcome mode collapse when dealing with discrete data, Fedus \emph{et al.}~\shortcite{fedus2018maskgan} proposed MaskGAN which used an actor-critic conditional GAN to fill in missing text conditioned on the surrounding context. Along this line, Wang \emph{et al.}~\shortcite{wang2018sentigan} proposed SentiGAN to generate texts of different sentiment labels. Besides, Li et al. \shortcite{li2018learning} improved the performance of semi-supervised text classification using adversarial training, \cite{zeng2018adversarial,qin2018dsgan} designed GAN-based models for distance supervision relation extraction.

Although various GAN based approaches have been explored for many applications, none of these approaches tackles open-domain event extraction from online texts. We propose a novel GAN-based event extraction model called AEM. Compared with the previous models, AEM has the following differences: (1) Unlike most GAN-based text generation approaches, a generator network is employed in AEM to learn the projection function between an event distribution and the event-related word distributions (entity, location, keyword, date). The learned generator captures event-related patterns rather than generating text sequence; (2) Different from LEM and DPEMM, AEM uses a generator network to capture the event-related patterns and is able to mine events from different text sources (short and long). Moreover, unlike traditional inference procedure, such as Gibbs sampling used in LEM and DPEMM, AEM could extract the events more efficiently due to the CUDA acceleration; (3) The discriminative features learned by the discriminator of AEM provide a straightforward way to visualize the extracted events. 

\section{Methodology}

We describe Adversarial-neural Event Model (AEM) in this section. An event is represented as a quadruple \textless$e,l,k,d$\textgreater, where $e$ stands for non-location named entities, $l$ for a location, $k$ for event-related keywords, $d$ for a date, and each component in the tuple is represented by component-specific representative words. 

AEM is constituted by three components: (1) \emph{The document representation module}, as shown at the top of Figure~\ref{fig:aem_framework}, defines a document representation approach which converts an input document from the online text corpus into $\vec d_{r}\in\mathbb{R}^{V}$ which captures the key event elements; (2) \emph{The generator} $G$, as shown in the lower-left part of Figure\ref{fig:aem_framework}, generates a fake document $\vec d_{f}$ which is constituted by four multinomial distributions using an event distribution $\vec \theta$ drawn from a Dirichlet distribution as input; (3) \emph{The discriminator} $D$, as shown in the lower-right part of Figure\ref{fig:aem_framework}, distinguishes the real documents from the fake ones and its output is subsequently employed as a learning signal to update the $G$ and $D$. The details of each component are presented below.

\subsection{Document Representation}

Each document {\color{black}$doc$} in a given corpus $C$ is represented as a concatenation of 4 multinomial distributions which are entity distribution ($\vec d_{r}^{e}$), location distribution ($\vec d_{r}^{l}$), keyword distribution ($\vec d_{r}^{k}$) and date distribution ($\vec d_{r}^{d}$) of the document. As four distributions are calculated in a similar way, we only describe the computation of the entity distribution below as an example.

The entity distribution $\vec d_{r}^{e}$ is represented by a normalized $V_{e}$-dimensional vector weighted by TF-IDF, and it is calculated as:
\begin{align*}
tf_{i,doc}^{e} & =\frac{n_{i,doc}^{e}}{\sum_{v_{e}}n_{v_{e},doc}^{e}} \\
idf_{i}^{e} &=\log \frac{|C^{e}|}{|C^{e}_{i}|}\\
tf\textrm{-}idf&_{i,doc}^{e} =tf_{i,doc}^{e}\times idf_{i}^{e} \\
d_{r,i}^{e} =&\frac{tf\textrm{-}idf_{i,doc}^{e}}{\sum_{v_{e}}tf\textrm{-}idf^{e}_{v_{e},doc}}
\end{align*}
where $C^{e}$ is the pseudo corpus constructed by removing all non-entity words from $C$, $V_{e}$ is the total number of distinct entities in a corpus, $n_{i,doc}^{e}$  denotes the number of $i$-th entity appeared in document $doc$, $|C^{e}|$ represents the number of documents in the corpus, and $|C_{i}^{e}|$ is the number of documents that contain $i$-th entity, and the obtained $d_{r,i}^{e}$ denotes the relevance between $i$-th entity and document $doc$. 

Similarly, location distribution $\vec d_{r}^{l}$, keyword distribution $\vec d_{r}^{k} $ and date distribution $\vec d_{r}^{d}$ of $doc$ could be calculated in the same way, and the dimensions of these distributions are denoted as $V_{l}$, $V_{k}$ and $V_{d}$, respectively. Finally, each document $doc$ in the corpus is represented by a $V$-dimensional ($V$=$V_{e}$+$V_{l}$+$V_{k}$+$V_{d}$) vector $\vec d_{r}$ by concatenating four computed distributions. 

\subsection{Network Architecture}

\subsubsection{Generator}

The generator network $G$ is designed to learn the projection function between the document-event distribution $\vec \theta$ and the four {\color{black}document-level word distributions} (entity distribution, location distribution, keyword distribution and date distribution). 

More concretely, $G$ consists of a $E$-dimensional document-event distribution layer, $H$-dimensional hidden layer and $V$-dimensional event-related word distribution layer. {\color{black}Here, $E$ denotes the event number, $H$ is the number of units in the hidden layer, $V$ is the vocabulary size and equals to $V_{e}$+$V_{l}$+$V_{k}$+$V_{d}$.}  As shown in Figure~\ref{fig:aem_framework}, $G$ firstly employs a random document-event distribution $\vec \theta$ as an input. To model the multinomial property of the document-event distribution, $\vec \theta$ is drawn from a Dirichlet distribution parameterized with $\vec \alpha$ which is formulated as:
\begin{align}
p(\vec\theta|\vec \alpha)= &Dir(\vec\theta|\vec\alpha)\triangleq\frac{1}{\triangle\left(\vec\alpha\right)}\prod_{t=1}^{E}\theta_{t}^{\alpha_{t}-1}
\end{align}

where $\vec \alpha$ is the hyper-parameter of the dirichlet distribution, $E$ is the number of events which should be set in AEM, $\triangle(\vec\alpha)=\frac{\prod_{t=1}^{E}\Gamma(\alpha_{t})}{\Gamma(\sum_{t=1}^{E}\alpha_{t})}$, $\theta_{t}\in [0,1]$ represents the proportion of event $t$ in the document and  $\sum_{t=1}^{E}\theta_{t}=1$.

Subsequently, $G$ transforms $\vec \theta$ into a $H$-dimensional hidden space using a linear layer followed by layer normalization, and the transformation is defined as:
{\color{black}\begin{align}
\vec s_{h}&=LN(W_{h}\vec  \theta+\vec b_{h}) \\
\vec o_{h}&=\max (\vec s_{h},l_{p}\times\vec s_{h})
\end{align}}
where $W_{h}\in\mathbb{R}^{H\times E}$ represents the weight matrix of hidden layer, and $\vec b_{h}$ denotes the bias term, $l_{p}$ is the parameter of LeakyReLU activation and is set to 0.1, $\vec s_{h}$ and $\vec o_{h}$ denote the normalized hidden states and the outputs of the hidden layer, and $LN$ represents the layer normalization. 

Then, to project $\vec o_{h}$ into four document-level event related word distributions ($\vec d_{f}^{e}$, $\vec d_{f}^{l}$, $\vec d_{f}^{k}$ and $\vec d_{f}^{d}$ shown in Figure \ref{fig:aem_framework}), four subnets (each contains a linear layer, a batch normalization layer and a softmax layer) are employed in $G$. And the exact transformation is based on the formulas below:
\begin{align}
\vec h_{w}^{e}=W_{w}^{e}\vec o_{h}+\vec b_{w}^{e},\  \vec{d_{f}^{e}}  =SM(BN(\vec h_{w}^{e}))\\
\vec h_{w}^{l}=W_{w}^{l}\vec o_{h}+\vec b_{w}^{l},\  \vec{d_{f}^{l}} =SM(BN(\vec h_{w}^{l}))\\
\vec h_{w}^{k}=W_{w}^{k}\vec o_{h}+\vec b_{w}^{k},\  \vec{d_{f}^{k}}=SM(BN(\vec h_{w}^{k}))\\
\vec h_{w}^{d}=W_{w}^{d}\vec o_{h}+\vec b_{w}^{d},\  \vec{d_{f}^{d}}=SM(BN(\vec h_{w}^{d}))
\end{align}
where $SM$ means softmax layer, $W_{w}^{e}\in \mathbb{R}^{V_{e}\times H}$, $W_{w}^{l}\in \mathbb{R}^{V_{l}\times H}$, $W_{w}^{k}\in \mathbb{R}^{V_{k}\times H}$ and $W_{w}^{d}\in \mathbb{R}^{V_{d}\times H}$ denote the weight matrices of the linear layers in subnets, $\vec b_{w}^{e}$, $\vec b_{w}^{l}$, $\vec b_{w}^{k}$ and $\vec b_{w}^{d}$ represent the corresponding bias terms, $\vec h_{w}^{e}$, $\vec h_{w}^{l}$, $\vec h_{w}^{k}$ and $\vec h_{w}^{d}$ are state vectors. $\vec d_{f}^{e}$, $\vec d_{f}^{l}$, $\vec d_{f}^{k}$ and $\vec d_{f}^{d}$ denote the generated entity distribution, location distribution, keyword distribution and date distribution, respectively, that correspond to the given event distribution $\vec \theta$. And each dimension represents the relevance between corresponding entity/location/keyword/date term and the input event distribution. 

Finally, four generated distributions are concatenated to represent the generated document $\vec d_{f}$ corresponding to the input $\vec \theta$:
\begin{equation}
\vec d_{f}=[\vec d_{f}^{e}; \vec d_{f}^{l};\vec d_{f}^{k};\vec d_{f}^{d}]
\end{equation}

\subsubsection{Discriminator}

The discriminator network $D$ is designed as a fully-connected network which contains an input layer, a discriminative feature layer (discriminative features are employed for event visualization) and an output layer. In AEM, $D$ uses fake document $\vec d_{f}$ and real document $\vec d_{r}$ as input and outputs the signal $D_{out}$ to indicate the source of the input data (lower value denotes that $D$ is prone to predict the input data as a fake document and vice versa). 

As have previously been discussed in \cite{arjovsky2017wasserstein,gulrajani2017improved}, lipschitz continuity of $D$ network is crucial to the training of the GAN-based approaches. To ensure the lipschitz continuity of $D$, we employ the spectral normalization technique \cite{miyato2018spectral}. More concretely, for each linear layer $l_{d}(\vec h)=W\vec h$ (bias term is omitted for simplicity) in $D$, the weight matrix $W$ is normalized by $\sigma(W)$. Here, $\sigma(W)$ is the spectral norm of the weight matrix $W$ with the definition below:
\begin{align}
\sigma(W):=\max\limits_{\vec h: \vec h \ne \vec 0} \frac{\|W\vec h\|_{2}}{\|\vec h\|_{2}}=\max \limits_{\|\vec h\|_{2}\le 1}\|W\vec h\|_{2}
\end{align}
which is equivalent to the largest singular value of $W$. The weight matrix $W$ is then normalized using:
\begin{equation}
\hat{W_{SN}}:=W/\sigma(W)
\end{equation}
Obviously, the normalized weight matrix $\hat{W_{SN}}$ satisfies that $\sigma(\hat{W_{SN}})=1$ and further ensures the lipschitz continuity of the $D$ network \cite{miyato2018spectral}. To reduce the high cost of computing spectral norm $\sigma(W)$ using singular value decomposition at each iteration, we follow \cite{yoshida2017spectral} and employ the power iteration method to estimate $\sigma(W)$ instead. With this substitution, the spectral norm can be estimated with very small additional computational time. 

\subsection{Objective and Training Procedure}

The real document $\vec d_{r}$ and fake document $\vec d_{f}$ shown in Figure~\ref{fig:aem_framework} could be viewed as random samples from two distributions $\mathbb{P}_{r}$ and $\mathbb{P}_{g}$, and each of them is a joint distribution constituted by four Dirichlet distributions (corresponding to entity distribution, location distribution, keyword distribution and date distribution). The training objective of AEM is to let the distribution $\mathbb{P}_{g}$ (produced by $G$ network) to approximate the real data distribution $\mathbb{P}_{r}$ as much as possible. 

To compare the different GAN losses, Kurach \shortcite{kurach2018gan} takes a sober view of the current state of GAN and suggests that the Jansen-Shannon divergence used in \cite{goodfellow2014generative} performs more stable than variant objectives. Besides, Kurach also advocates that the gradient penalty (GP) regularization devised in \cite{gulrajani2017improved} will further improve the stability of the model. Thus, the objective function of the proposed AEM is defined as:
\begin{align}
L_{d} &=-\underset{\vec d_{r}\sim \mathbb{P}_{r}}{\mathbb{E}}[\log(D(\vec d_{r}))]- \underset{\vec d_{f}\sim \mathbb{P}_{g}}{\mathbb{E}}[\log (1-D(\vec d_{f}))]\\
&\quad L_{gp}= \underset{ \vec{d^{*}}\sim\mathbb{P}_{d^{*}}}{\mathbb{E}}[(\parallel\nabla_{\vec{d^{*}}}D(\vec{d^{*}})\parallel_2 -1)^{2}]\\
&\quad\quad\quad\quad L = L_{d}+\lambda L_{gp}
\end{align}
where $L_{d}$ denotes the discriminator loss, $L_{gp}$ represents the gradient penalty regularization loss, $\lambda$ is the gradient penalty coefficient which trade-off the two components of objective, $\vec d^{*}$ could be obtained by sampling uniformly along a straight line between $\vec d_{r}$ and $\vec d_{f}$, $\mathbb{P}_{d^{*}}$ denotes the corresponding distribution.

The training procedure of AEM is presented in {\color{black}Algorithm~\ref{alg:1}, where $E$ is the  event number, $n_{d}$ denotes the number of discriminator iterations per generator iteration, $m$ is the batch size, $\alpha'$ represents the learning rate, $\beta_{1}$ and $\beta_{2}$ are hyper-parameters of Adam \cite{kingma2014adam}, $p_{a}$ denotes $\{\alpha',\beta_{1},\beta_{2}\}$.} {\color{black} In this paper, we set $\lambda=10$, $n_{d}=5$, $m=32$. Moreover, $\alpha'$, $\beta_{1}$ and $\beta_{2}$ are set as {\color{black}0.0002, 0.5 and 0.999}.

\begin{algorithm}[!h]
	\renewcommand{\algorithmicrequire}{\textbf{Input:}}
	\renewcommand{\algorithmicensure}{\textbf{Output:}}
	\caption{Training procedure for AEM}
	\label{alg:1}
	\begin{algorithmic}[1]
		\REQUIRE $E$, $\lambda$, $n_{d}$, $m$, $\alpha'$, $\beta_{1}$, $\beta_{2}$
		\ENSURE the trained $G$ and $D$.
		\STATE Initial $D$ parameters $\omega_{d}$ and $G$ parameter $\omega_{g}$
		\WHILE{$\omega_{g}$ has not converged}
		\FOR{$t=1,...,n_{d}$}
		\FOR{$j=1,...,m$}
		\STATE Sample $\vec d_{r}\sim \mathbb{P}_{r}$, 
		\STATE Sample a random  $\vec\theta\sim Dir(\vec\theta|\vec\alpha)$ 
		\STATE Sample a random number $\epsilon\sim U[0,1]$
		\STATE $\vec d_{f}\leftarrow G(\vec\theta)$
		\STATE $\vec{d^{*}}\leftarrow \epsilon \vec d_{r}+(1-\epsilon) \vec d_{f}$
		\STATE $L_{d}^{(j)}=-\log[D(\vec d_{r})]-\log [1-D(\vec d_{f})]$
		\STATE $L_{gp}^{(j)}=(\parallel \nabla_{\vec{d^{*}}}D(\vec{d^{*}}) \parallel-1)^{2}$
		\STATE $L^{(j)}\leftarrow L_{d}^{(j)}+\lambda L_{gp}^{(j)}$
		\ENDFOR
		\STATE $\omega_{d}\leftarrow Adam(\nabla_{\omega_{d}}\frac{1}{m}\sum\limits_{j=1}^{m}L^{(j)},\omega_{d},p_{a}) $
		\ENDFOR
		\STATE Sample $m$ noise $\left\{ \vec\theta^{(j)}\sim Dir(\vec\theta|\vec\alpha) \right\}$
		\STATE $\omega_{g}\leftarrow Adam(\nabla_{\omega_{g}}\frac{1}{m}\sum\limits_{j=1}^{m}\log [1-D(G(\vec\theta^{(j)}))],\omega_{g},p_{a})$
		\ENDWHILE 	
	\end{algorithmic}
\end{algorithm}

\subsection{Event Generation}

{\color{black}After the model training, the generator $G$ learns the mapping function between the document-event distribution and the document-level event-related word distributions (entity, location, keyword and date).} In other words, with an event distribution $\vec \theta'$ as input, $G$ could generate the corresponding entity distribution, location distribution,  keyword distribution and date distribution. 


In AEM, we employ event seed $\vec{s}_{t\in \{1,...,E\}}$, an $E$-dimensional vector with one-hot encoding, to generate the event related word distributions. For example, in ten event setting, $\vec{s}_{1}=[1,0,0,0,0,0,0,0,0,0]^{\intercal}$ represents the event seed of the first event. With the event seed $\vec s_{1}$ as input, the corresponding distributions could be generated by $G$ based on the equation below:
\begin{align}
[\vec \phi_{e}^{1};\vec \phi_{l}^{1};\vec \phi_{k}^{1};\vec \phi_{d}^{1}]=G(\vec s_{1})
\end{align}
where $\vec \phi_{e}^{1}$, $\vec \phi_{l}^{1}$, $\vec \phi_{k}^{1}$ and $\vec \phi_{d}^{1}$ denote the entity distribution, location distribution, keyword distribution and date distribution of the first event respectively.

\section{Experiments}

\begin{figure*}[!htbp]
\centering
\includegraphics[
  width=1.0\textwidth,
  keepaspectratio]
{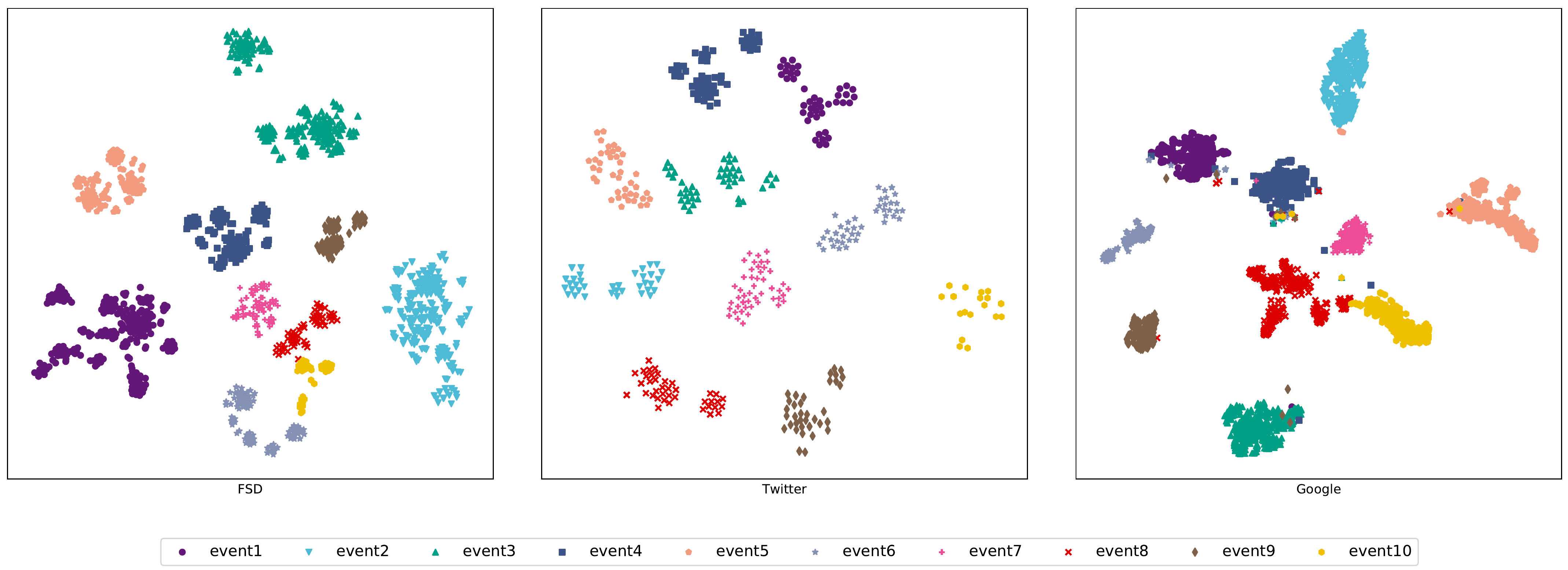}
\caption{{\color{black}Visualization of the ten randomly selected events on each dataset. Each point denotes a document. Different color denotes different events.}}
\label{fig:visualization}
\end{figure*}

\begin{figure*}[!ht]
\centering
\includegraphics[
  width=1.0\textwidth,
  keepaspectratio]
{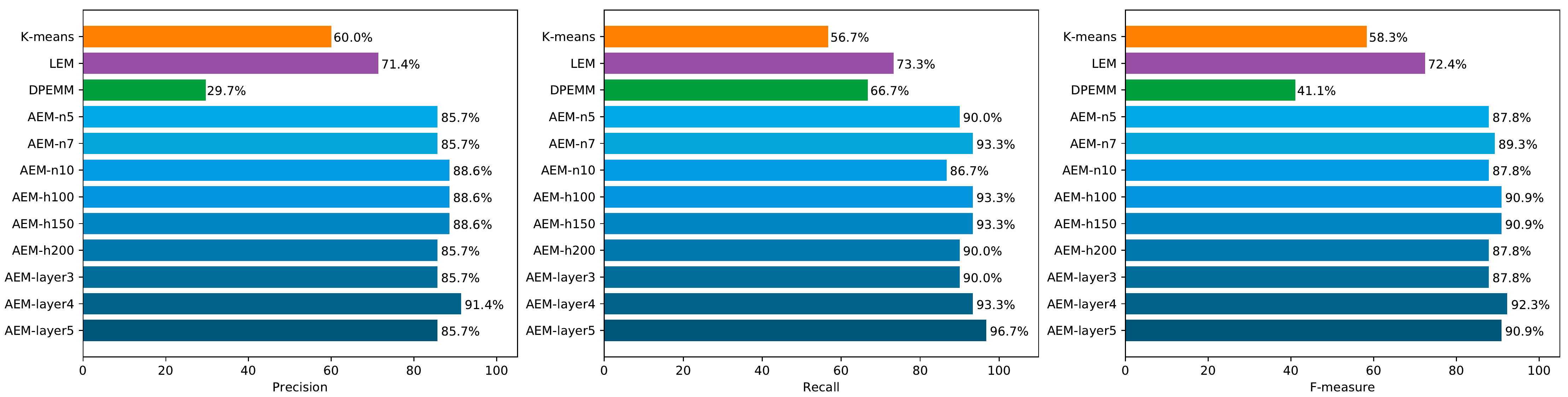}
\caption{Comparison of methods and parameter settings,{\color{black} `n' and `h' denote parameter $n_{d}$ and $H$, 
other parameters follow the default setting. The vertical axis represents methods/parameter settings, the horizontal axis denotes the corresponding performance value. All blue histograms with different intensity are those obtained by AEM. }}
\label{fig:para_cmp}
\end{figure*}

In this section, we firstly describe the datasets and baseline approaches used in our experiments and then present the experimental results. 

\subsection{Experimental Setup}

To validate the effectiveness of AEM for extracting events from social media (e.g. Twitter) and news media sites (e.g. Google news), three datasets (FSD \cite{petrovic2013can}, Twitter, and Google datasets\footnote{http://data.gdeltproject.org/events/index.html \label{d3}}) are employed. Details are summarized below:\\
\begin{itemize}
\item \underline{\emph{FSD dataset}} (social media) is the first story detection dataset containing 2,499 tweets. We filter out events mentioned in less than 15 tweets since events mentioned in very few tweets are less likely to be significant. The final dataset contains 2,453 tweets annotated with 20 events. \\
\item \underline{\emph{{\color{black}Twitter} dataset}} (social media) is collected from tweets published in the month of December in 2010 using Twitter streaming API. It contains 1,000 tweets annotated with 20 events. \\
\item \underline{\emph{Google dataset}} (news article) is a subset of GDELT Event Database$^{\ref{d3}}$, documents are retrieved by event related words. For example, documents which contain `malaysia', `airline', `search' and `plane' are retrieved for event \emph{MH370}. By combining 30 events related documents, the dataset contains 11,909 news articles.\\
\end{itemize}
We choose the following three models as the baselines:\\
\begin{itemize}
\item \underline{\textbf{K-means}} is a well known data clustering algorithm, we implement the algorithm using sklearn\footnote{https://scikit-learn.org/} toolbox, and represent documents using bag-of-words weighted by TF-IDF. \\
\item \underline{\textbf{LEM}} \cite{zhou2014simple} is a Bayesian modeling approach for open-domain event extraction. It treats an event as a latent variable and models the generation of an event as a joint distribution of its individual event elements. We implement the algorithm with the default configuration. \\
\item \underline{\textbf{DPEMM}} \cite{zhou2017event} is a non-parametric mixture model for event extraction. It addresses the limitation of LEM that the number of events should be known beforehand. We implement the model with the default configuration.
\end{itemize}

For social media text corpus (FSD and Twitter), a named entity tagger\footnote{http://fithub.com/aritter/twitter-nlp} specifically built for Twitter is used to extract named entities including locations from tweets. A Twitter Part-of-Speech (POS) tagger \cite{gimpel2010part} is used for POS tagging and only words tagged with nouns, verbs  and adjectives are retained as keywords. For the Google dataset, we use the Stanford Named Entity Recognizer\footnote{https://nlp.stanford.edu/software/CRF-NER.html} to identify the named entities  (organization, location and person). Due to the `date' information not being provided in the Google dataset, we further divide the non-location named entities into two categories (`person' and `organization') and employ a quadruple \textless organization, location, person, keyword\textgreater\ to denote an event in news articles. We also remove common stopwords and only keep the recognized named entities and the tokens which are verbs, nouns or adjectives. 

\subsection{Experimental Results}

To evaluate the performance of the proposed approach, we use the evaluation metrics such as precision, recall and F-measure. Precision is defined as the proportion of the correctly identified events out of the model generated events. Recall is defined as the proportion of correctly identified true events. For calculating the precision of the 4-tuple, we use following criteria:
\begin{itemize}
\item 
(1) Do the entity/organization, location, date/person and keyword that we have extracted refer to the same event?
\item 
(2) If the extracted representation contains keywords, are they informative enough to tell us what happened?
\end{itemize}

\begin{table}[!htbp]
\centering
\scalebox{0.67}{
\begin{tabular}{l|c|ccc}
\hline
{\bfseries Dataset}& {\bfseries Method}&{\bfseries Precision (\%)}&{\bfseries Recall (\%)}&{\bfseries F-measure (\%)} \\
\hline
\multirow{4}*{FSD}&K-means &84.0 &55.0 &66.5\\
&LEM & 80.0&80.0 &80.0\\
&DPEMM &84.6 &85.0 &84.8\\
&AEM & \textbf{88.0}& \textbf{85.0} &\textbf{86.5}\\
\hline
\multirow{4}*{Twitter}&K-means &68.0 &75.0 &71.3\\
&LEM &68.0 &80.0 &73.5\\
&DPEMM &69.2 &80.0 &74.2\\
&AEM &\textbf{72.0} &\textbf{85.0} &\textbf{77.9}\\
\hline
\multirow{4}*{Google}&K-Means &60.0 &56.7 &58.3\\
&LEM &71.4 &73.3 &72.4\\
&DPEMM &29.7 &66.7 &41.3\\
&AEM &\textbf{85.7} &\textbf{90.0} &\textbf{87.8}\\
\hline
\end{tabular}}
\caption{Comparison of the performance of event extraction on the three datasets.}
\label{tbs:results} 
\end{table}

Table~\ref{tbs:results} shows the event extraction results on the three datasets. The statistics are obtained with the default parameter setting that $n_{d}$ is set to 5, number of hidden units $H$ is set to 200, and $G$ contains three fully-connected layers. {\color{black}The event number $E$ for three datasets are set to 25, 25 and 35, respectively. The examples of extracted events are shown in Table.\ref{tbs:example_event}.

\begin{table*}[!h]
\scalebox{0.65}{
\begin{tabular}{c|l|c|l}
\hline
\multicolumn{4}{c}{{\bfseries FSD dataset}}\\
\hline
\multirow{4}{*}{\makecell[cc]{Earthquake \\ in Viriginia}}&e: nbc coast tremor east eastern&\multirow{4}{*}{\makecell[cc]{US \\ debt ceiling}}& e: hous gifford us gabriell repres\\
&l: virginia russian eal croydon washington&&l: virginia russian eal croydon washington\\
&k: earthquak feel center magnitud hit&& k: debt bill hous ceil vote\\
&d: 2011/8/23 2011/7/23 2011/8/06 2011/9/07 2011/9/12&&d: 2011/8/01 2011/7/23 2011/8/23 2011/8/06 2011/9/13\\
\hline
\multirow{4}{*}{\makecell[cc]{South sudan\\ independent}}&e: south sudan independ earthquak tremor&\multirow{4}{*}{\makecell[cc]{US \\ credit downgrade}}& e: aaa aa yahoo standard obama\\
&l: earth senat congress york nyc&&l: state america tottenham congress seattl\\
&k: independ celebr countri congrat challeng&& k: credit rate downgrad histori lose\\
&d: 2011/7/09 2011/8/06 2011/8/23 2011/7/23 2011/9/07&&d: 2011/8/06 2011/7/23 2011/8/23 2011/9/07 2011/9/12\\
\hline

\multirow{4}{*}{\makecell[cc]{Somalia \\ declare famine}}&e: somalia africa bakool southern nation&\multirow{4}{*}{\makecell[cc]{Norway youth \\ camp attack}}&e: eyewit norway norweigan rock us\\
&l: somalia africa rome independ southern&&l: norway island germani state libya\\
&k: declar famin drought part region&&k: camp attack youth bomb shoot\\
&d: 2011/7/20 2011/7/23 2011/8/06 2011/8/23 2011/9/07&&d: 2011/7/22 2011/7/23 2011/8/23 2011/8/06 2011/8/10 \\
\hline
\multicolumn{4}{c}{{\bfseries Twitter dataset}}\\
\hline
\multirow{4}{*}{\makecell[cc]{Russia hosts \\ world cup}}&e: world cup william russia sport&\multirow{4}{*}{\makecell[cc]{Larry King's \\ last show}}&e: king larri cnn red vega\\
&l: qatar russia china europ beij&&l: uk state richardson unit south\\
&k: host cup reaction world triumph&&k: final show broadcast night year\\
&d: 2010/9/3 2010/9/10 2010/9/9  2010/9/8 2010/9/17&&d: 2010/9/17 2010/9/10 2010/9/8 2010/9/9 2010/9/26\\
\hline
\multirow{4}{*}{\makecell[cc]{Coach Urban \\ Meyer step down}}&e: meyer urban reid florida gator&\multirow{4}{*}{\makecell[cc]{Boxer floyd \\ Maweath is arrested}}&e: boxer floyd mayweath vega obama\\
&l: florida univers senat europ hous&&l: vega las beij europ itali\\
&k: coach step univers footbal accord&&k: guard boxer secur assault arrest\\
&d: 2010/9/8 2010/9/10 2010/9/9 2010/9/18 2010/9/17&&d: 2010/9/17 2010/9/9 2010/9/18 2010/9/8 2010/9/26\\
\hline
\multirow{4}{*}{\makecell[cc]{Christian violence\\ in nigeria}}&e: christian muslim polit concord eve&\multirow{4}{*}{\makecell[cc]{Xiaobo Liu \\award nobel prize}}&e: xiaobo liu nobel prize china\\
&l: nigeria jos congress christian of&&l: china oslo congress continent europ\\
&k: religion church violenc plagu peopl&&k: award live nobel ceremoni dissid\\
&d: 2010/9/25 2010/9/28 2010/9/26 2010/9/6 2010/9/8&&d: 2010/9/10 2010/9/8 2010/9/17 2010/9/9 2010/9/18\\
\hline
\multicolumn{4}{c}{{\bfseries Google dataset}}\\
\hline

\multirow{4}{*}{\makecell[cc]{Sexual assault\\ in campus}}&o: university federal college department white&\multirow{4}{*}{\makecell[cc]{Lockett is executed\\ death penalty\\ in Oklahoma}}&o: warner state department cohen robert\\
&l: obama princeton ohio columbia harvard&&l: lockett oklahoma states texas ohio\\
&p: mccaskill rose catherine brown duncan&&p: lockett clayton patton stephanie charles\\
&k: sexual assault campus title colleges&&k: execution death penalty lethal minutes\\
\hline
\multirow{4}{*}{\makecell[cc]{Apple \& Samsung\\ patent jury}}&o: apple samsung google inc motorola&\multirow{4}{*}{\makecell[cc]{MH370}}&o: airlines air transport boeing najib\\
&l: california south santa us calif &&l: malaysia australia beijing malacca houston\\
&p: judge steve dunham schmidt mueller&&p: najib hishammuddin hussein clark dolan\\
&k: patent jury smartphone verdict trial&&k: search plane flight aircraft ocean\\
\hline
\multirow{4}{*}{\makecell[cc]{Afghanistan\\ landslide}}&o: afghanistan united taliban kabul un&\multirow{4}{*}{\makecell[cc]{South Africa\\ election}}&o: anc national mandela congress eff\\
&l: afghanistan badakhshan kabul tajikistan pakistan &&l: zuma africa south africans nkandla\\
&p: karzai shah hill mark angela&&p: zuma jacob president nelson malema\\
&k: landslide village rescue mud province&&k: election apartheid elections voters economic\\
\hline\end{tabular}
}
\caption{The event examples extracted by AEM. }
\label{tbs:example_event}
\end{table*}

It can be observed that K-means performs the worst over all three datasets. On the social media datasets, AEM outpoerforms both LEM and DPEMM by 6.5\% and 1.7\% respectively in F-measure on the FSD dataset, and 4.4\% and 3.7\% in F-measure on the Twitter dataset. We can also observe that apart from K-means, all the approaches perform worse on the Twitter dataset compared to FSD, possibly due to the limited size of the Twitter dataset. {\color{black}Moreover, on the Google dataset, the proposed AEM performs significantly better than LEM and DPEMM. It improves upon LEM by 15.5\% and upon DPEMM by more than 30\% in F-measure. This is because: (1) the assumption made by LEM and DPEMM that all words in a document are generated from a single event is not suitable for long text such as news articles; (2) DPEMM generates too many irrelevant events which leads to a very low precision score.} Overall, we see the superior performance of AEM across all datasets, with more significant improvement on the for Google datasets (long text). 

{\color{black}We next visualize the detected events based on the discriminative features learned by the trained $D$ network in AEM. The t-SNE \cite{maaten2008visualizing} visualization results on the datasets are shown in Figure~\ref{fig:visualization}}. For clarity, each subplot is plotted on a subset of the dataset containing ten randomly selected events. {\color{black} It can be observed that documents describing the same event have been grouped into the same cluster.} 

{\color{black}To further evaluate if a variation of the parameters $n_{d}$ (the number of discriminator iterations per generator iteration), $H$ (the number of units in hidden layer) and the structure of generator $G$ will impact the extraction performance, additional experiments have been conducted on the Google dataset, with $n_{d}$ set to 5, 7 and 10, $H$ set to 100, 150 and 200, and three $G$ structures (3, 4 and 5 layers). The comparison results on precision, recall and F-measure are shown in Figure~\ref{fig:para_cmp}. {\color{black} From the results, it could be observed that AEM with the 5-layer generator performs the best and achieves 96.7\% in F-measure, and the worst F-measure obtained by AEM is 85.7\%.} {\color{black}Overall, the AEM outperforms all compared approaches acorss various parameter settings, showing relatively stable performance.}

Finally, we compare in Figure~\ref{fig:time_cmp} the training time required for each model, excluding the constant time required by each model to load the data. We could observe that K-means runs fastest among all four approaches. Both LEM and DPEMM need to sample the event allocation for each document and update the relevant counts during Gibbs sampling which are time consuming. AEM only requires a fraction of the training time compared to LEM and DPEMM. {\color{black}Moreover, on a larger dataset such as the Google dataset, AEM appears to be far more efficient compared to LEM and DPEMM.}
\begin{figure}[!htbp]
\centering
\includegraphics[
  width=0.5\textwidth,
  keepaspectratio]
{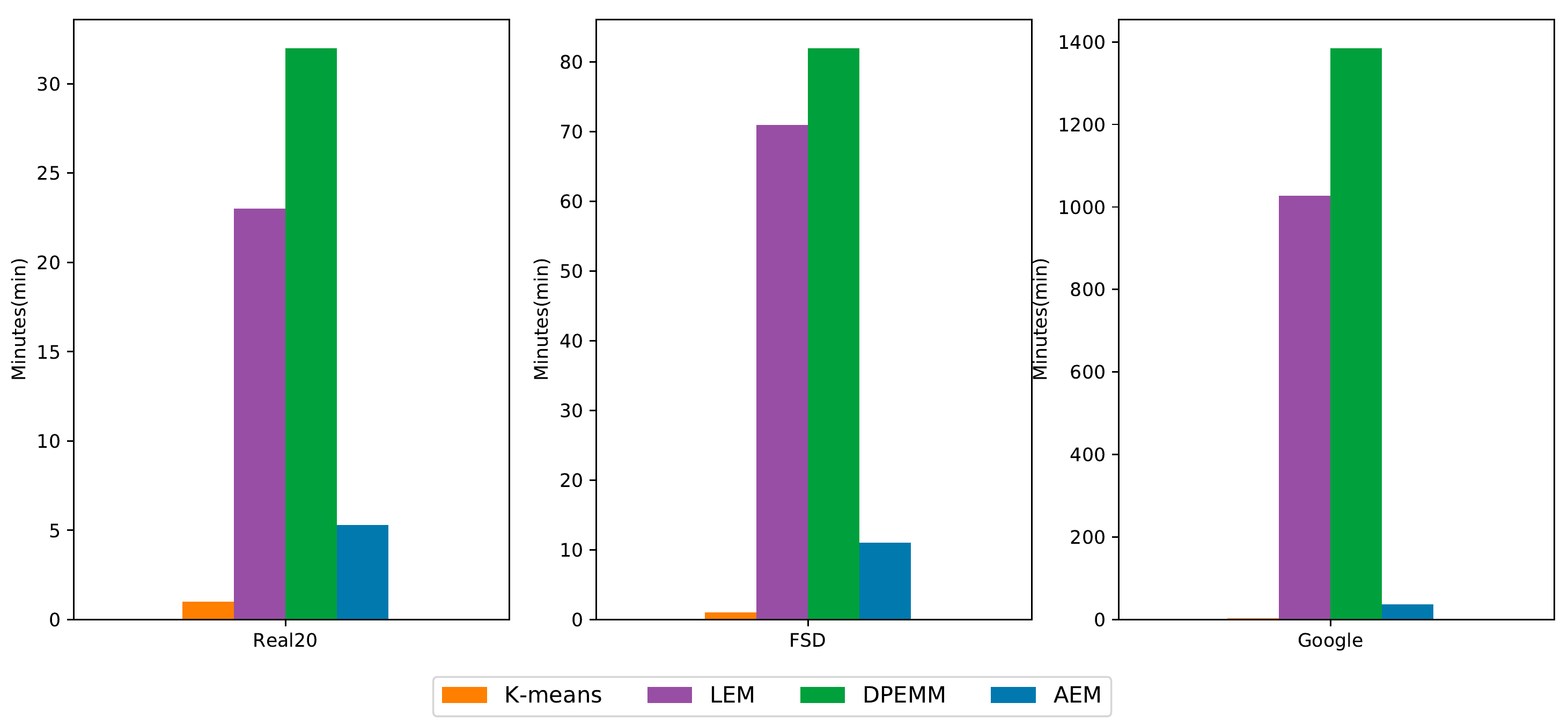}
\caption{{\color{black}Comparison of training time of models.}}
\label{fig:time_cmp}
\end{figure}
\section{Conclusions and Future Work}
In this paper, we have proposed a novel approach based on adversarial training to extract the structured representation of events from online text. 
The experimental comparison with the state-of-the-art methods shows that AEM achieves improved extraction performance, especially on long text corpora with an improvement of 15\% observed in F-measure. AEM only requires a fraction of training time compared to existing Bayesian graphical modeling approaches. 
In future work, we will explore incorporating external knowledge (e.g. word relatedness contained in word embeddings) into the learning framework for event extraction. Besides, exploring nonparametric neural event extraction approaches and detecting the evolution of events over time from news articles are other promising future directions. 

\section{Acknowledgments}
We would like to thank anonymous reviewers for their valuable comments and helpful suggestions. This work was funded by the National Key Research and Development Program of China (2016YFC1306704), the National Natural Science Foundation of China (61772132), the Natural Science Foundation of Jiangsu Province of China (BK20161430).

\bibliographystyle{acl_natbib}

\end{document}